\documentclass{article}
\usepackage{spconf,amsmath,graphicx,hyperref}
\usepackage{multirow}
\usepackage{xcolor}
\usepackage{booktabs, graphicx}
\usepackage[nameinlink]{cleveref}
\crefname{figure}{Fig.}{Figs.}
\Crefname{figure}{Fig.}{Figs.}
\usepackage{amssymb}  
\usepackage{amsfonts}
\usepackage{sectsty}
\usepackage{enumitem}
\subsectionfont{\fontsize{10}{5}\selectfont\bfseries}
\usepackage{subcaption}
\usepackage[T1]{fontenc}
\usepackage{cleveref}

\crefname{table}{Table}{Tables}
\Crefname{table}{Table}{Tables}

\title{IQ-LUT: Interpolated and Quantized LUT for Efficient Image Super-Resolution}

\name{Yuxuan Zhang$^{1}$, Zhikai Dong$^{2}$, Xinning Chai$^{1}$, Xiangyun Zhou$^{2}$, Yi Xu$^{1}$, Zhengxue Cheng$^{1,\dagger}$\thanks{$^{\dagger}$Corresponding author.},
Li Song$^{1,\dagger}$ }

\address{$^{1}$ Shanghai Jiao Tong University \\
      $^{2}$ Rockchip Electronics Co., Ltd \\
      {\tt\small \{67keudyhsi,chaixinning,xuyi,zxcheng,song\_li\}@sjtu.edu.cn} \\ 
     {\tt\small\{zhikai.dong,zxy\}@rock-chips.com}
}
     
\begin{document}
%
\maketitle
\begin{abstract}

Lookup table (LUT) methods demonstrate considerable potential in accelerating image super-resolution inference. However, pursuing higher image quality through larger receptive fields and bit-depth triggers exponential growth in the LUT's index space, creating a storage bottleneck that limits deployment on resource-constrained devices. We introduce \textbf{IQ-LUT}, which achieves a reduction in LUT size while simultaneously enhancing super-resolution quality. 
First, we integrate interpolation and quantization into the single-input, multiple-output ECNN, which dramatically reduces the index space and thereby the overall LUT size. 
Second, the integration of residual learning mitigates the dependence on LUT bit-depth, which facilitates training stability and prioritizes the reconstruction of fine-grained details for superior visual quality. 
Finally, guided by knowledge distillation, our non-uniform quantization process optimizes the quantization levels, thereby reducing storage while also compensating for quantization loss. 
Extensive benchmarking demonstrates our approach substantially reduces storage costs (by up to \textbf{50$\times$} compared to ECNN) while achieving superior super-resolution quality.

\end{abstract}

\begin{keywords}
Interpolation, Residual learning, Non-uniform quantization, Knowledge distillation, Lookup table
\end{keywords}

\vspace{-2mm}
\section{Introduction}
\label{sec:intro}
\vspace{-2mm}
With the growing demand for real-time high-quality image restoration on mobile devices and embedded platforms, lightweight, low-latency single image super-resolution (SISR) methods have become a research hotspot. Lookup table (LUT)-based acceleration is an effective method for resource-constrained devices because it pre-computes the mapping from low-resolution to high-resolution image patches and replaces online inference with efficient table indexing.

\begin{figure}[t] 
  \centering
  \begin{subfigure}[b]{0.36\columnwidth}
  \includegraphics[width=\linewidth]{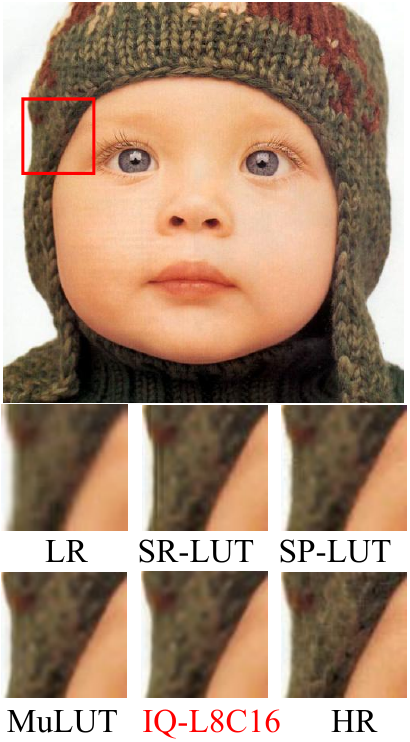}
\end{subfigure}
\hspace{0.02\columnwidth}
\begin{subfigure}[b]{0.47\columnwidth}
  \includegraphics[width=\linewidth]{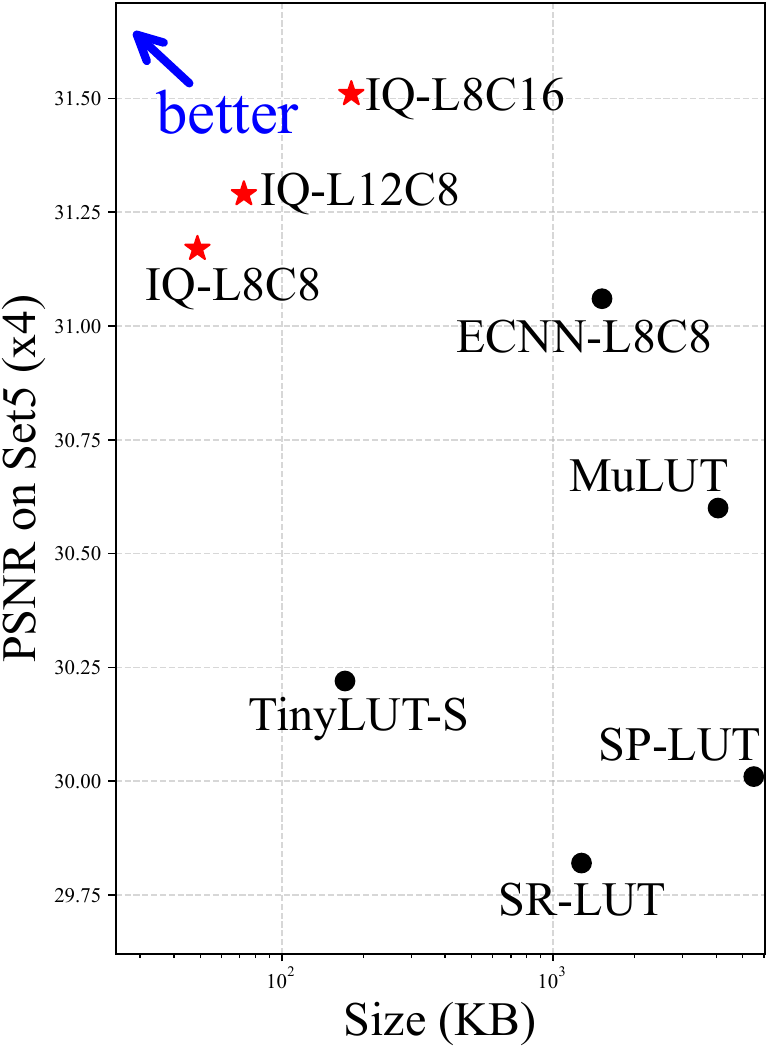}
\end{subfigure}

  \caption{Qualitative and quantitative comparison on Set5 for x4 SR. As can be seen from the left figure, our model IQ-L8C16 performs well on the boundary. As can be seen from the right figure, our three IQ-LUT models all achieve small LUT-Size and high PSNR.}
  \vspace{-4mm}
  \label{fig:visual_results}
\end{figure}

LUT-based SR methods, such as SR-LUT\cite{srlut}, usually improve quality by enlarging the receptive field, which leads to an increase in the LUT index range and an exponential growth in LUT size. Subsequent studies\cite{auto,eclut,imlut,user,xidian,tiny,spf}, such as MuLUT\cite{MuLUT}, adopted strategies such as interpolation, rotation, and compact encoding to replace storage with additional computation, thereby achieving a balance between accuracy and efficiency. However, these methods still suffer from the problems of large overall model size and low quality.
\begin{figure*}[t] 
    \centering
    \includegraphics[width=\textwidth]{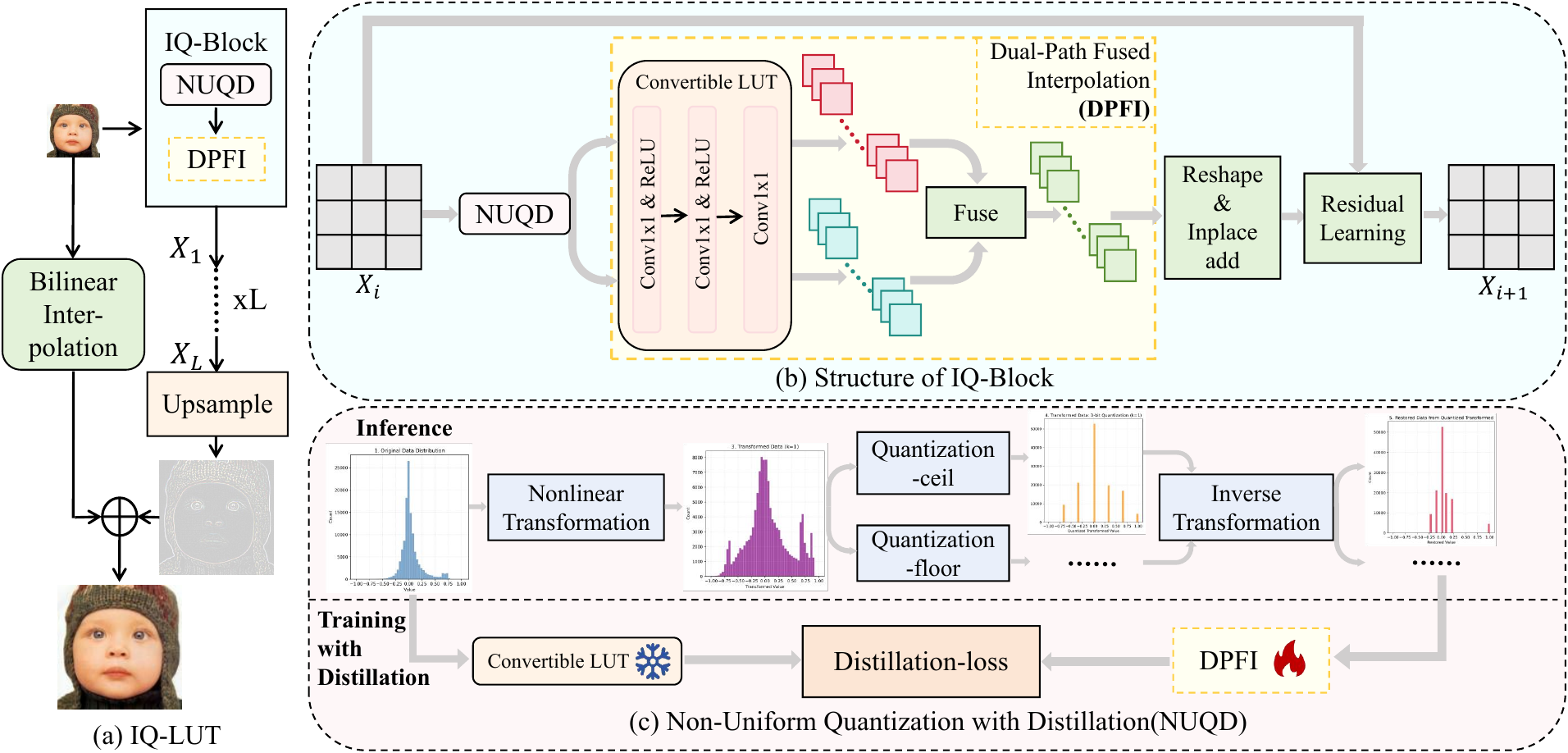} 
    \caption{The overview of our model.(a) represents the overall structure of IQ-LUT, (b) is the internal structure of IQ-Block, which is the core of our model. It consists of two parts: NUQD and DPFI. (c) is the internal structure of NUQD. The lower part is the distillation training process, the snowflakes on the left represent freezing, and the flames on the right represent trainable.}
     \vspace{-3mm}
    \label{fig:overview}
\end{figure*}

Compared with the aforementioned multi-input LUT method, ECNN\cite{ecnn} introduces a novel expanded convolution that maps a single pixel to multiple output values, thus achieving a better balance between size and quality. However, the LUT size still grows exponentially with increasing index bit-depth. To address this, we propose IQ-LUT, an expanded convolution-based model that makes a hardware-efficient trade-off: it introduces minimal computation to circumvent the prohibitive cost of exponential storage growth, a strategy that benefits dedicated hardware where memory dominates.
First, in order to reduce the size, we adopt bilinear interpolation to establish a low-frequency foundation and solely train the residual component, so that the network focuses on high frequencies and the output distribution is more concentrated. Learnable residual connections are incorporated within each IQ-Block to facilitate stable gradient propagation, enable the training of deeper and wider network architectures, and enhance overall SR performance.
Secondly, simply reducing the bit-depth also leads to severe quality degradation. Therefore, in each IQ-Block, we propose an interpolation scheme, called Dual-Path Fused Interpolation (DPFI), which adopts low index bit-depth and interpolates intermediate values instead of explicitly storing them. This strategy enables significant model compression with minimal loss in visual quality. 
Finally, training only the residuals yields a more concentrated output distribution. This concentration renders uniform quantization suboptimal, as it inefficiently allocates storage to value ranges with low occupancy. To address this issue, We add a non-uniform quantization module (NUQD) at the input of each  IQ-Block, applying piecewise-linear mapping followed by quantization to compress redundant regions. We also use a high bit-depth model as a teacher model for knowledge distillation. This reduces the LUT size by eliminating redundancy while achieving finer discretization in important regions, thereby improving SR quality.

The specific contributions of our IQ-LUT are as follows: 
\begin{itemize}
\setlist{nosep}
\item We propose an IQ-LUT comprising stacked IQ-Blocks to explicitly learn high-frequency residuals, which significantly improves detail recovery in image super-resolution.
\vspace{-2mm}
\item We propose Dual Path Fusion Interpolation (DPFI), which reduces the input bit-depth while replacing explicit storage with interpolation, effectively balancing model size and reconstruction quality.
\vspace{-2mm}
\item We propose Non-uniform Quantization with Distillation (NUDQ), which employs piecewise mapping for finer discretization of key regions and leverages knowledge distillation to improve quality.
 \vspace{-1mm}
\end{itemize}

\vspace{-2mm}
\section{PROPOSED METHOD}

\label{sec:method}
\vspace{-1mm}
\subsection*{\textit{A.Preliminary}}
\vspace{-1mm}
\begin{table*}[t]
    \centering
    \scriptsize 
    \setlength{\tabcolsep}{4pt}
     \caption{Quantitative results of 
$\times4$ super-resolution on five benchmark datasets. The best results are highlighted in \textbf{bold}, while the second best are marked with underline. Size denotes the LUT storage size. }
\vspace{-3mm}
    \label{tab:x4_results}
    \vspace{2pt}
    \resizebox{\textwidth}{!}{
    \begin{tabular}{c c c c c c c}
        \toprule
        Model &  Size(KB) & Set5 & Set14 & BSD100 & Urban100 & Manga109 \\
        &  & (PSNR$\uparrow$ \ SSIM$\uparrow$) & (PSNR$\uparrow$ \ SSIM$\uparrow$) & (PSNR$\uparrow$ \ SSIM$\uparrow$) & (PSNR$\uparrow$ \ SSIM$\uparrow$) & (PSNR$\uparrow$ \ SSIM$\uparrow$) \\
        \midrule
        Nearest &  - & 26.25\quad 0.7372 & 24.65\quad 0.6529 & 25.03\quad 0.6293 & 22.17\quad 0.6154 & 23.45\quad 0.7414 \\
Bilinear  & - & 27.55\quad 0.7884 & 25.42\quad 0.6792 & 25.54\quad 0.6460 & 22.69\quad 0.6346 & 24.21\quad 0.7666 \\
Bicubic & - & 28.42\quad 0.8101 & 26.00\quad 0.7023 & 25.96\quad 0.6672 & 23.14\quad 0.6574 & 24.91\quad 0.7871 \\
        \midrule
        SR-LUT  & 1274    & 29.82\quad 0.8478 & 27.01\quad 0.7355 & 26.53\quad 0.6953 & 24.02\quad 0.6990 & 26.80\quad 0.8380 \\
        SP-LUT  & 5500    & 30.01\quad 0.8516 & 27.21\quad 0.7427 & 26.67\quad 0.7019 & 24.12\quad 0.7058 & 27.00\quad 0.8430 \\
        MuLUT & 4062     & 30.60\quad 0.8653 & 27.60\quad 0.7541 & 26.86\quad 0.7110 & 24.46\quad 0.7194 & 27.90\quad 0.8633 \\
    
        TinyLUT-F & 171  & 31.18\quad 0.8771 & \underline{28.01}\quad 0.7630 & 27.13\quad 0.7184 & 24.92\quad 0.7397 & 28.83\quad 0.8798 \\
        TinyLUT-S  & \underline{37}   & 30.22\quad 0.8535 & 27.33\quad 0.7450 & 26.71\quad 0.7042 & 24.19\quad 0.7066  & 27.21\quad 0.8458 \\

        ECNN-L8C8  & 1516  & 31.06\quad 0.8753 & 27.91\quad 0.7631 & 27.08\quad 0.7180 & 24.82\quad 0.7364 & 28.59\quad 0.8762 \\
        
        IQ-L8C8   & \textbf{34}  & 31.14\quad 0.8761 & 27.93\quad 0.7634 & 27.09\quad 0.7183 & 24.84\quad 0.7373 & 28.64\quad 0.8767 \\
        IQ-L12C8  & 50  & \underline{31.26}\quad \underline{0.8794} & 28.00\quad \underline{0.7660} & \underline{27.14}\quad \underline{0.7204} & \underline{24.96}\quad \underline{0.7427} & \underline{28.86}\quad \underline{0.8817} \\
        IQ-L8C16  & 124 & \textbf{31.50}\quad \textbf{0.8838} & \textbf{28.12}\quad \textbf{0.7697} & \textbf{27.22}\quad \textbf{0.7238} & \textbf{25.14}\quad \textbf{0.7500} & \textbf{29.17}\quad \textbf{0.8878} \\
        
        \bottomrule
    \end{tabular}
    }
   \vspace{-3mm}
\end{table*}

Our model is built on the expanded convolutional (EC) neural network (ECNN) with $L$ stacked EC layers, followed by an upsample module, which is implemented as a specialized EC layer integrated with a PixelShuffle operation. 
Each EC layer is designed as a lightweight subnetwork comprising three $1 \times 1$ convolutional layers and two ReLU activations, as illustrated in the Convertible LUT module in \cref{fig:overview}. During the training phase, this subnetwork generates intermediate features for each input pixel. At inference time, it is converted into a LUT to enhance computational efficiency:
\begin{equation}
X(i, j, c) = \Phi_\theta(F_{\text{in}}(i, j, c)),
\end{equation}
where $F_{\text{in}}$ is the input and $\Phi_\theta$ denotes the subnetwork. 
The final output $F_{n,c,h,w}$ is obtained by the "Reshape and
Inplace add" windows:
\begin{equation}
F_{n,c,h,w} = \sum_{i=0}^{k_h-1} \sum_{j=0}^{k_w-1} \sum_{c_{\text{in}}=0}^{C_{\text{in}}-1} 
X_{\text{patch}}[n, c_{\text{in}}, c, i, j, h+i, w+j],
\end{equation}
where $X_{\text{patch}}$ is obtained by rearranging $X(i, j, c)$.

\vspace{-2mm}
\subsection*{\textit{B.The whole structure of IQ-LUT}}
\vspace{-1mm}
As illustrated in \cref{fig:overview} (a), Our IQ-LUT consists of L layers of IQ-Blocks. Each IQ-Block, as shown in part (b) of the \cref{fig:overview}, sequentially undergoes non-uniform quantization (NUQD), dual-path fuse interpolation (DPFI), and a learnable residual connection. Finally, after L layers of IQ-Blocks, the output undergoes upsample, then summed with the bilinear interpolation of the low-resolution image to produce a high-resolution image.
It mitigates the network's reliance on high bit-depth. Furthermore, each IQ-Block incorporates a learnable scalar parameter $\alpha$ to connect the input residual to the output, facilitating adaptive information flow and enabling the training of deeper, wider networks:
\vspace{-1mm}
\begin{equation}
x_{\text{out}} = (1-\sigma(\alpha)) \cdot x + \sigma(\alpha) \cdot F(x),
\vspace{-1mm}
\end{equation}
where $\sigma(\cdot)$ denotes the sigmoid function. $F(x)$ is the result of $x$ after NUQD and DPFI module.

\vspace{-1mm}
\subsection*{\textit{C.NUDQ:Non-uniform quantization with distillation}}
\vspace{-1mm}
To address the trade-off between LUT size and reconstruction quality, we adopt non-uniform quantization to enhance bit-depth efficiency. Unlike standard uniform quantization within a fixed range, non-uniform quantization allows for finer discretization in more important regions, thereby reducing memory requirements while preserving key feature information.

Specifically, within each IQ-Block, the input is processed by a Non-uniform Quantization with Distillation (NUDQ) module. We introduce a symmetric piecewise-linear mapping \(T_{a,b}\) for its computational efficiency and hardware-friendly implementation:
\vspace{-1mm}
\begin{equation}
T_{a,b}(x)=
\begin{cases}
-1 + s_o(x+1), & x\le -a,\\[1pt]
s_m\,x,         & |x|<a,\\[1pt]
b + s_o(x-a),   & x\ge a,
\end{cases} \\
\vspace{-1mm}
\end{equation}
and \(s_m=\dfrac{b}{a},\ s_o=\dfrac{1-b}{1-a}, 0<a, b<1.\)
The hyperparameters $a$ and $b$ are optimized via a greedy search to obtain different slopes, which in turn facilitate distinct quantization effects.  Then We uniformly quantize and nonlinearly inverse transform $T_{a,b}(x)$.
To further stabilize the training and enhance the intermediate feature representation, as shown in \cref{fig:overview} (c), we also fine-tune the low bit-depth pre-trained student network with the high bit-depth pre-trained teacher network to complete knowledge distillation.

In our final model, the first IQ-Block use 4-bit input, while all subsequent blocks operate at 3-bit precision, with each IQ-Block producing 8-bit output. The distillation process is conducted from a 8-bit input, 12-bit output teacher network.

\vspace{-2mm}
\subsection*{\textit{D.DPFI:Dual-Path Fused
Interpolation}}
\vspace{-1mm}
A major challenge in LUT-based methods is that improving performance comes at the cost of an exponential increase in storage  size with higher bit-depth. However, A naive reduction of the bit-depth inevitably leads to a significant degradation in quality. To address this issue, we employ an interpolation scheme to approximate intermediate LUT values, thereby enhancing fidelity while preserving a low bit-depth.

As shown in \cref{fig:overview} (c),
NUQD quantizes an input by performing bidirectional rounding (both upward and downward), producing two outputs: $X_{\text{floor}}$ and $X_{\text{ceil}}$, which correspond to the nearest lower and upper LUT indices, respectively.
The interpolation weights $T$ are computed by:
\vspace{-1mm}
\[
T = (x_{\mathrm{trans}} - x_{\mathrm{floor}}) \cdot (2^{b-1} - 1), \quad T \in [0,1]
\]
\vspace{-1mm}
where $b$ denotes the target bit-depth and $x_{\mathrm{trans}}$ represents the output of nonlinear transformation in NUQD. 
The fused feature, which is the output of DPFI, is then obtained by a weighted combination:
\vspace{-1mm}
\begin{equation}
F(x) = (1 - T) \odot X_{\text{floor}} + T \odot X_{\text{ceil}}.
\vspace{-1mm}
\end{equation}
\vspace{-2mm}

\begin{figure}[t] 
    \centering
    \includegraphics[width=\columnwidth]{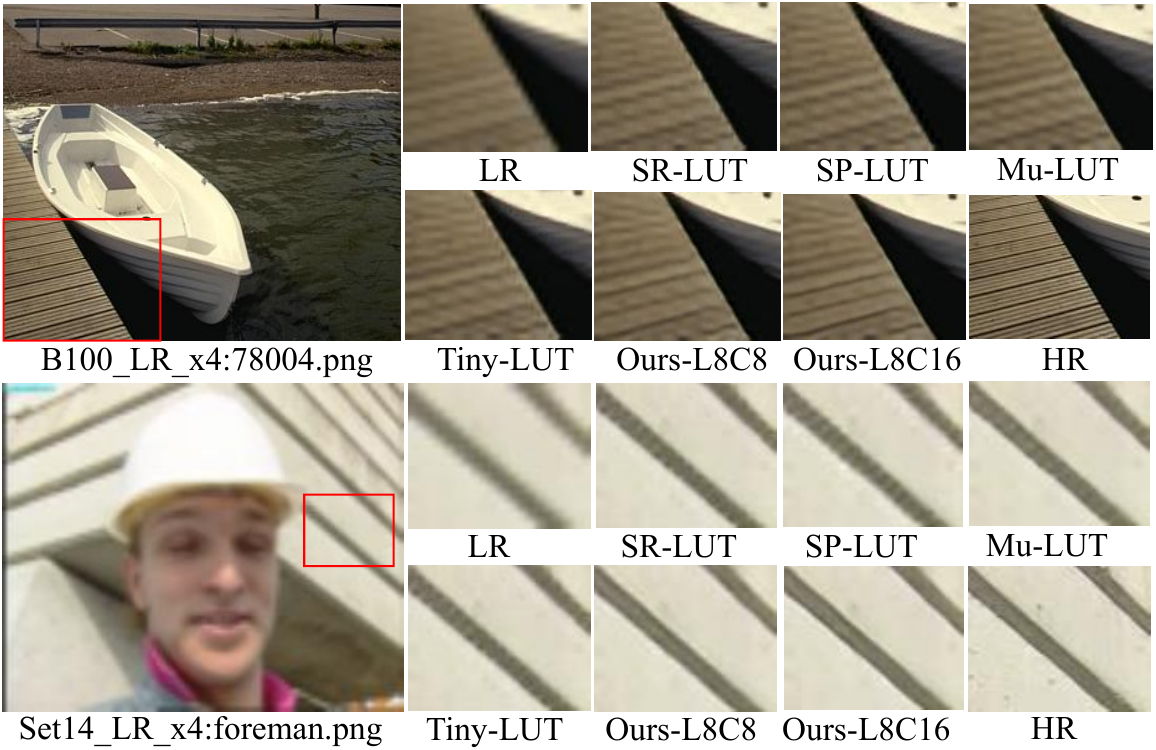} 
    \caption{Qualitative comparison of super-resolution results on Set14 and B100 using different LUT-based models and our IQ-LUT method.}
    \label{fig:visual_results}
    \vspace{-4mm}
\end{figure}

\vspace{-6mm}
\section{EXPERIMENTS}
\label{sec:exp}
\subsection*{\textit{A.Implementation Details}}
\vspace{-1mm}
\textbf{Datasets and Metrics} We employ the DIV2K dataset\cite{div2k} for training and evaluate on five standard benchmarks: Set5\cite{set5}, Set14\cite{set14}, B100\cite{b100}, Urban100\cite{urban100}, and Manga109\cite{manga109}. Quantitative performance is assessed using PSNR and SSIM computed on the Y channel in YCbCr space.

\noindent \textbf{Training Details}
The model is trained for $1\times10^6$ iterations using the Adam optimizer~\cite{adam} ($\beta_1=0.9$, $\beta_2=0.999$), with an initial learning rate of $1\times10^{-4}$ that is halved at 200K, 400K, 600K, and 800K iterations. The loss combines MSE (weight 1.0) and distillation loss (weight 3.0). Training consists of two stages: initial optimization with MSE for convergence, followed by fine-tuning with non-uniform quantization and distillation to reduce quantization effects. All experiments are implemented in PyTorch on an NVIDIA GeForce RTX 3090 GPU.

\vspace{-2mm}
\subsection*{\textit{B.Quantitative Comparison}}
\vspace{-1mm}
 As presented in \cref{tab:x4_results}, our model configurations, denoted IQ-LXCY, where X and Y correspond to the number of layers of IQ-Block and the number of channels of intermediate features, respectively, consistently surpass previous work across all benchmarks. IQ-L8C16 achieves the best PSNR and SSIM results on every dataset with only 124 KB. Even the compact IQ-L8C8 (34 KB) outperforms most LUT-based methods and larger models, demonstrating an excellent balance between efficiency and reconstruction quality and validating our design’s effectiveness.

\vspace{-2mm}
\subsection*{\textit{C.Qualitative Comparison}}
\vspace{-1mm}

Beyond quantitative metrics, we qualitatively evaluated the recovery of fine textures. As shown in \cref{fig:visual_results}, our IQ-LUT recovers sharper and more accurate textures than prior LUT-based methods. It better preserves complex structures and edges that are typically blurred or over-smoothed by previous methods. This demonstrates the efficacy of our DPFI and residual learning modules in reconstructing high-frequency details.

\vspace{-2mm}
\subsection*{\textit{D.Complexity Analysis}}
\vspace{-1mm}
As anticipated, the introduced interpolation incurs a modest latency overhead. Notably, our model delivers this performance at merely twice the latency of ECNN on GPU while requiring only $1/50$ of its parameters. This represents a highly favorable trade-off, exchanging minimal computation for a drastic reduction in storage footprint. Crucially, our primary objective is optimization for custom hardware (ASIC) deployment, where storage—not logic—dominates area and power costs. Consequently, our radical storage compression provides a decisive efficiency advantage that is not reflected in generic processor benchmarks.

\vspace{-2mm}
\subsection*{\textit{E.Ablation Study}}
\vspace{-1mm}
\textbf{a) The Effectiveness of DPFI and Residual Learning} 
To evaluate the contributions of the proposed components, we conduct ablation studies on five benchmark datasets using the IQ-L8C8 model. As summarized in \cref{tab:ablation1}, the DPFI module consistently improves PSNR, and further incorporating residual learning brings additional gains, confirming that both components are critical to enhancing reconstruction quality.

\begin{table}[t]
\centering
\caption{Effect of DPFI and Residual Learning (Res) Modules on PSNR Across Benchmark Datasets for $\times 4$ SR. The input quantization bit of every IQ-Block is 4.}
\vspace{-2mm}
\label{tab:ablation1}
\setlength{\tabcolsep}{4pt} 
\small 
\begin{tabular}{c|c|c c c c c}
\hline
DPFI & Res & Set5 & Set14 & B100 & Urban100 & Manga109 \\
\hline
& & 30.63 & 27.59 & 26.90 & 24.50 & 27.69 \\
\checkmark & & 31.04 & 27.89 & 27.06 & 24.75 & 28.44 \\
\checkmark & \checkmark & \textbf{31.20} & \textbf{27.99} & \textbf{27.13} & \textbf{24.91} & \textbf{28.74} \\
\hline
\end{tabular}
\vspace{-2mm}
\end{table}

\begin{table}[t]
\centering
\caption{Effect of NUQD on PSNR Across Benchmark Datasets for $\times 4$ SR. 
the first IQ-Block uses 4-bit input. All subsequent blocks operate use 3-bit input.}
\vspace{-2mm}
\small
\begin{tabular}{c|c|c|c|c|c}
\hline
NUQD & Set5 & Set14 & B100 & Urban100 & Manga109 \\
\hline
           & 31.12 & 27.91 & 27.09 & 24.82 & 28.52 \\
\checkmark & \textbf{31.17} & \textbf{27.95} & \textbf{27.10} & \textbf{24.85} & \textbf{28.65} \\
\hline
\end{tabular}
\label{tab:ablation_quant}
\vspace{-1mm}
\end{table}

\noindent
\textbf{b) The Impact of NUDQ} 
We also analyze the impact of NUQD using the IQ-L8C8 model. Results in \cref{tab:ablation_quant} show that introducing NUQD consistently improves performance across all datasets. The results validate its effectiveness in improving reconstruction quality.

\vspace{-2mm}
\section{CONCLUSIONS}
\label{sec:conclusion}
\vspace{-1mm}
Our IQ-LUT address the challenges of LUT-based super-resolution by introducing Residual Learning, Dual-Path Fused Interpolation and Non-Uniform Quantization with Distillation. The proposed IQ-LUT achieves state-of-the-art performance across all benchmark datasets, notably attaining a PSNR of 31.50 dB on Set5 with its optimal configuration (IQ-L8C16), while maintaining a compact model size of only 124 KB. These strategies effectively alleviate the LUT size explosion problem and improve super-resolution quality.

\section{ACKNOWLEDGEMENTS}
This work was partly supported by the NSFC(62431015, 62571317, 62501387), Science and Technology Commission of Shanghai Municipality No.25511106700, the Fundamental Research Funds for the Central Universities, Shanghai Key Laboratory of Digital Media Processing and Transmission under Grant 22DZ2229005, 111 project BP0719010.

\bibliographystyle{IEEEbib}
\bibliography{refs}

@INPROCEEDINGS{srlut,
  author={Jo, Younghyun and Joo Kim, Seon},
  booktitle={2021 IEEE/CVF Conference on Computer Vision and Pattern Recognition (CVPR)}, 
  title={Practical Single-Image Super-Resolution Using Look-Up Table}, 
  year={2021},
  volume={},
  number={},
  pages={691-700},
  doi={10.1109/CVPR46437.2021.00075}}

@inproceedings{
ecnn,
title={Expanded Convolutional Neural Network Based Look-Up Tables for High Efficient Single-Image Super-Resolution},
author={Kai Yin and Jie Shen},
booktitle={ACM Multimedia 2024},
year={2024},
url={https://openreview.net/forum?id=mRsa7615AA}
}

@InProceedings{MuLUT,
      author    = {Li, Jiacheng and Chen, Chang and Cheng, Zhen and Xiong, Zhiwei},
      title     = {{MuLUT}: Cooperating Multiple Look-Up Tables for Efficient Image Super-Resolution},
      booktitle = {ECCV},
      year      = {2022},
  }

@misc{auto,
      title={AutoLUT: LUT-Based Image Super-Resolution with Automatic Sampling and Adaptive Residual Learning}, 
      author={Yuheng Xu and Shijie Yang and Xin Liu and Jie Liu and Jie Tang and Gangshan Wu},
      year={2025},
      eprint={2503.01565},
      archivePrefix={arXiv},
      primaryClass={cs.CV},
      url={https://arxiv.org/abs/2503.01565}, 
}

@INPROCEEDINGS{spf,
  author={Li, Yinglong and Li, Jiacheng and Xiong, Zhiwei},
  booktitle={2024 IEEE/CVF Conference on Computer Vision and Pattern Recognition (CVPR)}, 
  title={Look-Up Table Compression for Efficient Image Restoration}, 
  year={2024},
  volume={},
  number={},
  pages={26016-26025},
  doi={10.1109/CVPR52733.2024.02458}}

@INPROCEEDINGS{tiny,
  author={Li, Yinglong and Li, Jiacheng and Xiong, Zhiwei},
  booktitle={2024 IEEE/CVF Conference on Computer Vision and Pattern Recognition (CVPR)}, 
  title={Look-Up Table Compression for Efficient Image Restoration}, 
  year={2024},
  volume={},
  number={},
  pages={26016-26025},
  doi={10.1109/CVPR52733.2024.02458}}

@inproceedings{eclut,
author = {Yin, Kai and Shen, Jie},
title = {Efficient look-up table from expanded convolutional network for accelerating image super-resolution},
year = {2024},
isbn = {978-1-57735-887-9},
publisher = {AAAI Press},
url = {https://doi.org/10.1609/aaai.v38i7.28495},
doi = {10.1609/aaai.v38i7.28495},
booktitle = {Proceedings of the Thirty-Eighth AAAI Conference on Artificial Intelligence and Thirty-Sixth Conference on Innovative Applications of Artificial Intelligence and Fourteenth Symposium on Educational Advances in Artificial Intelligence},
articleno = {747},
numpages = {9},
series = {AAAI'24/IAAI'24/EAAI'24}
}

@article{xidian, title={Multi-Frame Deformable Look-Up Table for Compressed Video Quality Enhancement}, volume={39}, url={https://ojs.aaai.org/index.php/AAAI/article/view/32351}, DOI={10.1609/aaai.v39i3.32351},  number={3}, journal={Proceedings of the AAAI Conference on Artificial Intelligence}, author={He, Gang and Quan, Guancheng and Wu, Chang and Wang, Shihao and Zhou, Dajiang and Li, Yunsong}, year={2025}, month={Apr.}, pages={3392-3400} }

@misc{imlut,
      title={IM-LUT: Interpolation Mixing Look-Up Tables for Image Super-Resolution}, 
      author={Sejin Park and Sangmin Lee and Kyong Hwan Jin and Seung-Won Jung},
      year={2025},
      eprint={2507.09923},
      archivePrefix={arXiv},
      primaryClass={eess.IV},
      url={https://arxiv.org/abs/2507.09923}, 
}

@INPROCEEDINGS{user,
  author={Zhao, Xin and Hu, Zhicheng and Chang, Liang},
  booktitle={2024 IEEE International Symposium on Circuits and Systems (ISCAS)}, 
  title={USR-LUT: A High-Efficient Universal Super Resolution Accelerator with Lookup Table}, 
  year={2024},
  volume={},
  number={},
  pages={1-5},
  doi={10.1109/ISCAS58744.2024.10558295}}

@InProceedings{div2k,
	author = {Agustsson, Eirikur and Timofte, Radu},
	title = {NTIRE 2017 Challenge on Single Image Super-Resolution: Dataset and Study},
	booktitle = {The IEEE Conference on Computer Vision and Pattern Recognition (CVPR) Workshops},
	month = {July},
	year = {2017}
}

@inproceedings{set5,
  title={Low-Complexity Single-Image Super-Resolution based on Nonnegative Neighbor Embedding},
  author={Bevilacqua, Marco and Roumy, Aline and Guillemot, Christine and Morel, Marie-Line Alberi},
  booktitle={British Machine Vision Conference (BMVC)},
  year={2012}
}

@InProceedings{set14,
    author="Zeyde, Roman
    and Elad, Michael
    and Protter, Matan",
    editor="Boissonnat, Jean-Daniel
    and Chenin, Patrick
    and Cohen, Albert
    and Gout, Christian
    and Lyche, Tom
    and Mazure, Marie-Laurence
    and Schumaker, Larry",
    title="On Single Image Scale-Up Using Sparse-Representations",
    booktitle="Curves and Surfaces",
    year="2012",
    publisher="Springer Berlin Heidelberg",
    address="Berlin, Heidelberg",
    pages="711--730",
    isbn="978-3-642-27413-8"
}

@INPROCEEDINGS{b100,
  author={Martin, D. and Fowlkes, C. and Tal, D. and Malik, J.},
  booktitle={Proceedings Eighth IEEE International Conference on Computer Vision. ICCV 2001}, 
  title={A database of human segmented natural images and its application to evaluating segmentation algorithms and measuring ecological statistics}, 
  year={2001},
  volume={2},
  number={},
  pages={416-423 vol.2},
  doi={10.1109/ICCV.2001.937655}
}

@INPROCEEDINGS{urban100,
  author={Huang, Jia-Bin and Singh, Abhishek and Ahuja, Narendra},
  booktitle={2015 IEEE Conference on Computer Vision and Pattern Recognition (CVPR)}, 
  title={Single image super-resolution from transformed self-exemplars}, 
  year={2015},
  volume={},
  number={},
  pages={5197-5206},
  doi={10.1109/CVPR.2015.7299156}}

@article{manga109,
  title={Sketch-based manga retrieval using manga109 dataset},
  author={Yusuke Matsui and Kota Ito and Yuji Aramaki and Azuma Fujimoto and Toru Ogawa and T. Yamasaki and Kiyoharu Aizawa},
  journal={Multimedia Tools and Applications},
  year={2015},
  volume={76},
  pages={21811 - 21838},
  url={https://api.semanticscholar.org/CorpusID:8887614}
}

@article{adam,
  title={Adam: A Method for Stochastic Optimization},
  author={Diederik P. Kingma and Jimmy Ba},
  journal={CoRR},
  year={2014},
  volume={abs/1412.6980},
  url={https://api.semanticscholar.org/CorpusID:6628106}
}

\end{document}